%% file: main.tex
\documentclass[11pt, a4paper, logo, copyright, nonumbering]{deepseek}
\usepackage[numbers, sort&compress]{natbib}

\usepackage{dblfloatfix}
\usepackage{ulem}
\usepackage{caption}
\usepackage{dramatist}
\usepackage{xspace}
\usepackage{pifont}
\usepackage{multirow}
\usepackage{tcolorbox}
\usepackage{xltabular}
\usepackage{longtable}
\usepackage{pdfpages}
\definecolor{linkcolor}{rgb}{0.956,0.298,0.235}
\definecolor{citecolor}{HTML}{1976D2}
\usepackage{tabularx}
\usepackage{hyperref}
\usepackage{graphicx}
\usepackage{subfigure}
\usepackage{amsmath}
\usepackage{amssymb}
\usepackage{array}
\usepackage{booktabs}
\usepackage{multirow}
\usepackage{arydshln}
\usepackage[utf8]{inputenc}
\hypersetup{
    colorlinks=true,
    linkcolor=citecolor,
    filecolor=magenta,
    urlcolor=cyan,
    citecolor=citecolor,
}

\usepackage{amsfonts}
\usepackage{amsmath}
\usepackage{amssymb}
\usepackage{lineno}

\usepackage[bottom]{footmisc}

\usepackage{CJKutf8}
\usepackage{subfigure}
\usepackage{setspace}

\usepackage{xspace}

\makeatletter
\def\@BTrule[#1]{
  \ifx\longtable\undefined
    \let\@BTswitch\@BTnormal
  \else\ifx\hline\LT@hline
    \nobreak
    \let\@BTswitch\@BLTrule
  \else
     \let\@BTswitch\@BTnormal
  \fi\fi
  \global\@thisrulewidth=#1\relax
  \ifnum\@thisruleclass=\tw@\vskip\@aboverulesep\else
  \ifnum\@lastruleclass=\z@\vskip\@aboverulesep\else
  \ifnum\@lastruleclass=\@ne\vskip\doublerulesep\fi\fi\fi
  \@BTswitch}
\makeatother

\addto\extrasenglish{

}

\reportnumber{001}

\renewcommand{\today}{}

\title{\centering \includegraphics[scale=0.05, bb=0 166 200 0]{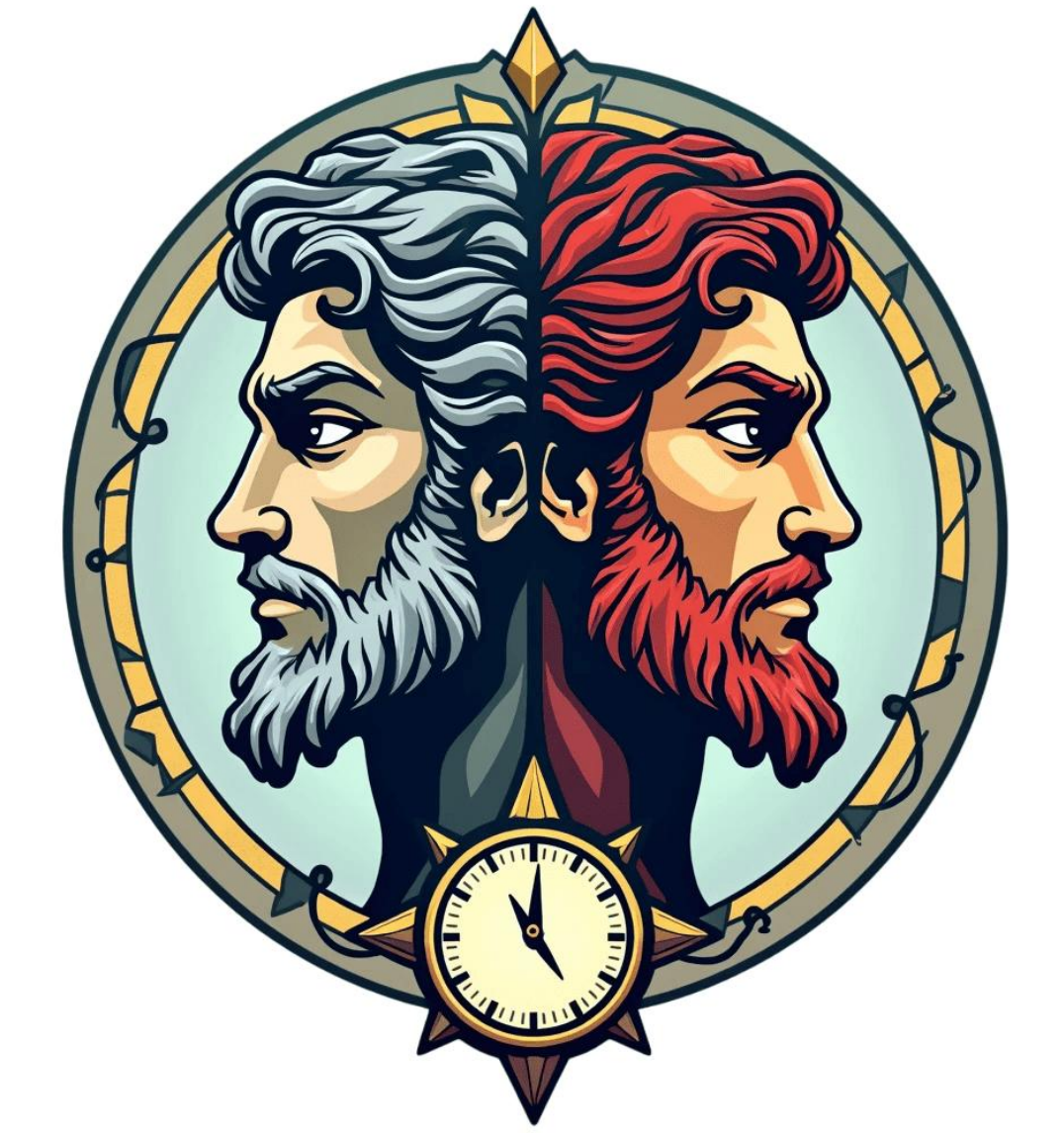} \quad \  Janus: Decoupling Visual Encoding for Unified Multimodal Understanding and Generation}

\author[*]{
\small

Chengyue Wu$^{1,2}$ \quad Xiaokang Chen$^{1, *, \dag}$ \quad Zhiyu Wu$^{1,3}$ \quad Yiyang Ma$^{1,3}$ \quad Xingchao Liu$^1$ \quad Zizheng Pan$^1$ \quad Wen Liu$^1$ \quad Zhenda Xie$^1$ \quad Xingkai Yu$^1$ \quad Chong Ruan$^1$ \quad Ping Luo$^{2, *}$ \\

\small
$^1$DeepSeek-AI \quad $^2$The University of Hong Kong \quad $^3$Peking University \\
\small
$^\dag$: Project lead \quad $^*$: Corresponding authors \\
\small
Project Page: \url{https://github.com/deepseek-ai/Janus}

}

\begin{abstract}
In this paper, we introduce \textbf{Janus}, an autoregressive framework that unifies multimodal understanding and generation. 
Prior research often relies on a single visual encoder for both tasks, such as Chameleon. However, due to the differing levels of information granularity required by multimodal understanding and generation, this approach can lead to suboptimal performance, particularly in multimodal understanding. To address this issue, we decouple visual encoding into separate pathways, while still leveraging a single, unified transformer architecture for processing.
The decoupling not only alleviates the conflict between the visual encoder's roles in understanding and generation, but also enhances the framework's flexibility. For instance, both the multimodal understanding and generation components can independently select their most suitable encoding methods.
Experiments show that Janus surpasses previous unified model and matches or exceeds the performance of task-specific models.
The simplicity, high flexibility, and effectiveness of Janus make it a strong candidate for next-generation unified multimodal models.

\end{abstract}

\begin{document}
\begin{CJK*}{UTF8}{gbsn}

\maketitle

\section{Introduction}

\begin{figure}[t]
\begin{center}
    \subfigure[Benchmark Performance.]
    {
        \includegraphics[width=0.41\linewidth]{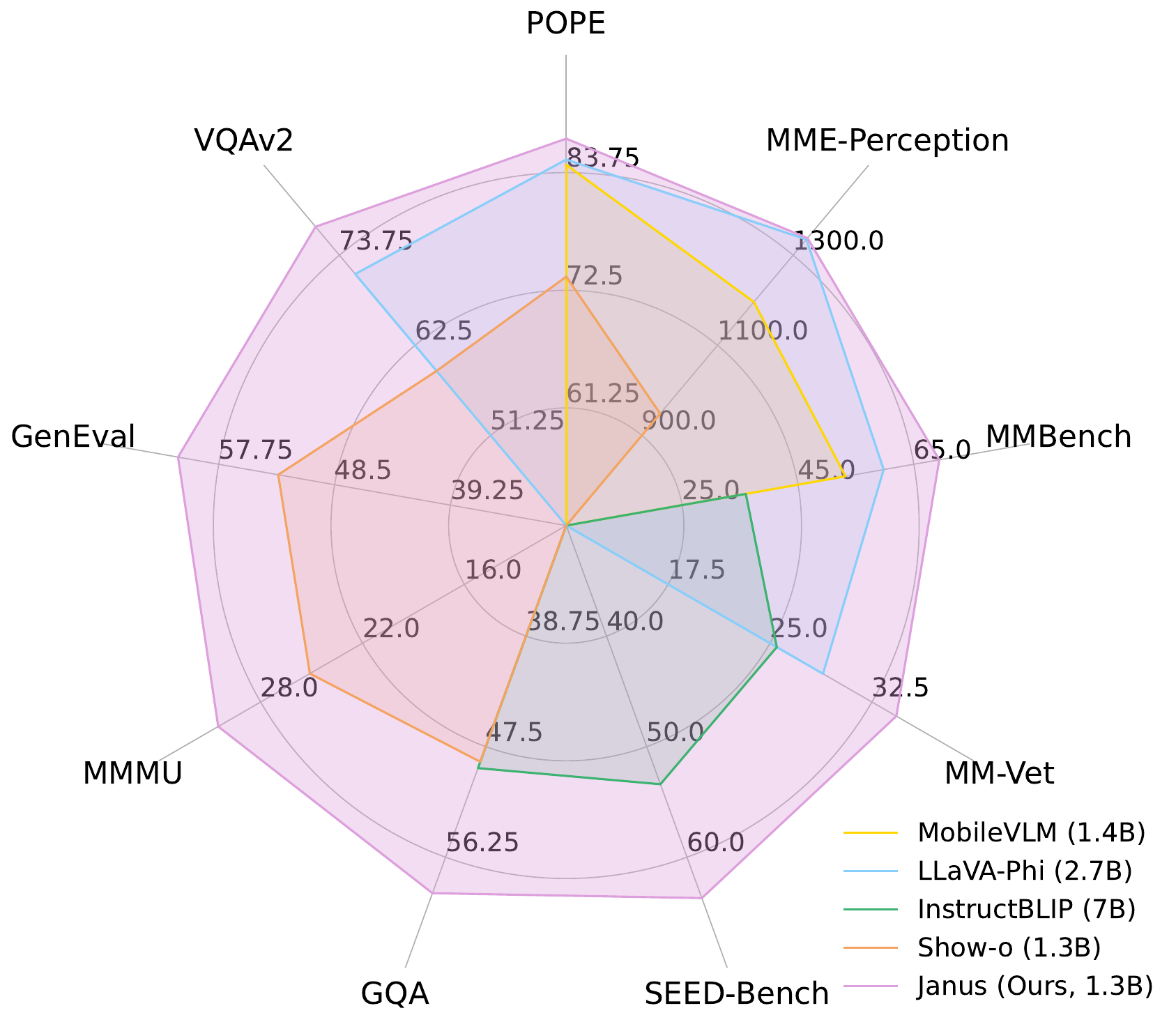}
        \label{fig:teaser_understanding}
    }
    \hfill
    \subfigure[Visual Generation Results.]
    {
        \includegraphics[width=0.55\linewidth]{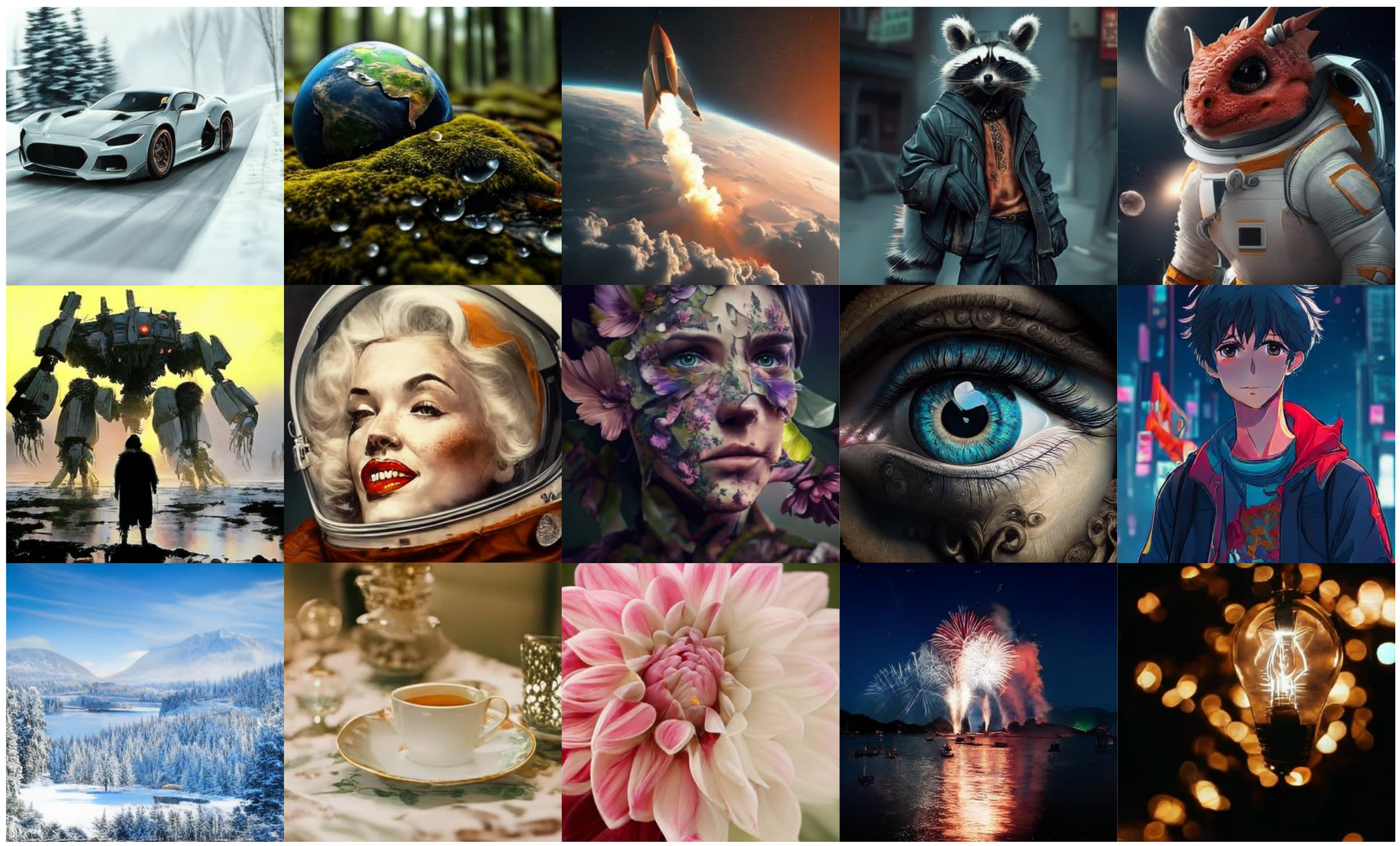}
        \label{fig:teaser_generation}
    }
\end{center}
\caption{\small
\textbf{Multimodal understanding and vision generation results from our Janus}. Janus outperforms the previous state-of-the-art unified multimodal models as well as some task-specific multimodal understanding models, while also demonstrating strong visual generation capabilities. The image resolution is $384 \times 384$. Best viewed on screen.
}
\vspace{-2pt}
\label{fig:teaser}
\end{figure}

In recent years, multimodal large models have made significant advancements in both understanding and generation domains~\cite{liu2024visual,dhariwal2021diffusion}. In the field of multimodal understanding, researchers follow the design of LLaVA~\cite{liu2024visual} by using a vision encoder as a bridge to enable large language models (LLMs) to understand images. In the field of visual generation, diffusion-based approaches~\cite{dhariwal2021diffusion,chen2023pixart,rombach2022high,dhariwal2021diffusion} have seen notable success. More recently, some works have explored autoregressive methods for vision generation~\cite{sun2024autoregressive,tian2024visual}, achieving performance comparable to diffusion models. To build more powerful and generalist multimodal models, researchers have sought to combine multimodal understanding and generation tasks~\cite{sun2023generative,team2024chameleon,zhou2024transfusion}. For instance, some studies have attempted to connect multimodal understanding models with pretrained diffusion models~\cite{sun2023generative,ge2023making,ge2023planting}. 
For example, Emu~\cite{sun2023generative} 
uses the output of the LLM as a condition for a pretrained diffusion model, and then relies on the diffusion model to generate images.
However, strictly speaking, this approach cannot be considered a truly unified model, because the visual generation functionality is handled by the external diffusion model, while the multimodal LLM itself lacks the capability to directly generate images.

Other approaches~\cite{team2024chameleon,wu2024vila,xie2024show,zhou2024transfusion} employ a single transformer to unify both multimodal understanding and generation tasks, which improves instruction-following for visual generation, unlocks potential emergent abilities, and reduces model redundancy.
Such methods typically use a single vision encoder to process inputs for both two tasks. However, the representations required by multimodal understanding and generation tasks differ significantly. 
In multimodal understanding tasks, the purpose of the vision encoder is to extract high-level semantic information (e.g., object categories or visual attributes within an image). The output of understanding task not only involves extracting information from images but also involves complex semantic reasoning. Therefore, the granularity of the vision encoder’s representation tends to mainly focus on high-dimensional semantic representation. By contrast, in visual generation tasks, the main focus is on generating local details and maintaining global consistency in the image. The representation in this context necessitates a low-dimensional encoding that is capable of fine-grained spatial structure and textural detail expression. Unifying the representations of these two tasks within the same space will lead to conflicts and trade-offs. Consequently, existing unified models for multimodal understanding and generation often compromise on multimodal understanding performance, falling markedly short of the state-of-the-arts multimodal understanding models. We explore this issue further in the ablation study.

To solve this problem, we propose \textbf{Janus}\footnote{In Roman mythology, Janus is the god of duality and transitions, symbolizing the coexistence of contradictory forces by having two faces, each looking in opposite directions. Similarly, our model captures the inherent tension between vision tasks: understanding demands abstract, high-level semantic representations, while generation requires concrete, detailed information. By decoupling these processes into specialized encoders, our system mirrors Janus's dual nature, resolving this tension within a unified architecture.}, a unified multimodal framework that decouples visual encoding for multimodal understanding and generation. Specifically, we introduce two independent visual encoding pathways: one for multimodal understanding and one for multimodal generation, unified by the same transformer architecture. The proposed method offers two main benefits: (1) Janus alleviates the conflict stemming from the different granular needs of multimodal understanding and generation and eliminates the need to make trade-offs between two tasks when selecting visual encoders. (2) Janus is flexible and extensible. After decoupling, both the understanding and generation tasks can adopt state-of-the-art encoding techniques specific to their domain. Moreover, it is possible for Janus to accommodate additional input types in the future, such as point clouds, EEG signals, or audio data, where independent encoders can extract features and then use a unified transformer to process them. 

To the best of our knowledge, we are the first to highlight the importance of decoupling visual encoding within the unified multimodal understanding and generation framework.
Our experimental results show that Janus surpasses existing unified models with comparable parameter sizes on both multimodal understanding and generation benchmarks, achieving state-of-the-art results. Notably, Janus even outperforms some task-specific models which have significantly more parameters (Figure~\ref{fig:teaser}). Specifically, on multimodal understanding benchmarks MMBench~\cite{liu2023mmbench}, SEED-Bench~\cite{li2023seed}, and POPE~\cite{li2023evaluating}, Janus ($1.3$B) achieved scores of $69.4$, $63.7$, and $87.0$, respectively, outperforming LLaVA-v$1.5$ ($7$B)~\cite{liu2024improved} and Qwen-VL-Chat ($7$B)~\cite{bai2023qwen} . On visual generation benchmarks MSCOCO-$30$K~\cite{chen2015microsoft} and GenEval~\cite{ghosh2024geneval}, Janus achieved an FID score of $8.53$ and an accuracy of $61$\%, surpassing text-to-image generative models such as DALL-E $2$~\cite{ramesh2022hierarchical} and SDXL~\cite{podell2023sdxl}. We believe that the strong performance, coupled with the high flexibility and extensibility of Janus, presents it as a strong candidate for next-generation unified multimodal models.

\input{sec/related_work}

\input{sec/method}
\input{sec/experiments}
\input{sec/conclusion}

\clearpage

\input{sec/appendix}

\clearpage

\bibliographystyle{abbrvnat}
\bibliography{main}

\end{CJK*}
\end{document}

%% file: sec/related_work.tex
\section{Related Work}

\subsection{Visual Generation}

Visual generation is a rapidly evolving field that combines concepts from natural language processing with advancements in transformer architectures. Autoregressive models, influenced by the success in language processing, leverage transformers to predict sequences of discrete visual tokens (codebook IDs)~\cite{esser2021taming, ramesh2021zero,sun2023generative}. These models tokenize visual data and employ a prediction approach similar to GPT-style~\cite{radford2018improving} techniques. Additionally, masked prediction models~\cite{chang2022maskgit, chang2023muse} draw upon BERT-style~\cite{devlin2018bert} masking methods, predicting masked sections of visual inputs to improve synthesis efficiency, and have been adapted for video generation~\cite{yu2023magvit}. Concurrently, continuous diffusion models have showcased impressive capabilities in visual generation~\cite{ho2020denoising, song2020denoising, rombach2022high}, complementing discrete methods by approaching generation through a probabilistic lens.

\subsection{Multimodal Understanding}
Multimodal large language models (MLLMs) integrate both text and images~\cite{brown2020language, touvron2023llama, touvron2023llama2}. By leveraging pretrained LLMs, MLLMs ~\cite{lu2024deepseek,liu2024visual,zhu2023minigpt,wang2024visionllm,chen2024far,Anthropic_Claude3,achiam2023gpt} demonstrate a robust ability to understand and process multimodal information. Recent advancements have explored extending MLLMs with pretrained diffusion models to facilitate image generation~\cite{ge2023planting,jin2023unified,sun2023generative, sun2024generative, ge2024seed}. These methods fall under the category of tool utilization, where diffusion models are used to generate images based on the conditions output by the MLLM, while the MLLM itself does not have the ability to directly perform visual generation. Moreover, the generative ability of the entire system is often constrained by the external diffusion model, making its performance inferior to directly using the diffusion model on its own~\cite{ge2023planting,sun2023generative}.

\subsection{Unified Multimodal Understanding and Generation}

Unified multimodal understanding and generation models are considered powerful for facilitating seamless reasoning and generation across different modalities \cite{team2024chameleon,zhou2024transfusion}. Traditional approaches in these models typically use a single visual representation for both understanding and generation tasks, regardless of whether they are based on autoregressive (AR) models \cite{team2024chameleon,wu2024vila} or diffusion models \cite{xie2024show,zhou2024transfusion}. 
For example, Chameleon~\cite{team2024chameleon} adopts a VQ Tokenizer to encode images for both multimodal understanding and generation.
However, this practice may lead to suboptimal outcomes, as the vision encoder might face a trade-off between the demands of understanding and generation. In contrast, our Janus can explicitly decouple the visual representations for understanding and generation, recognizing that different tasks may require varying levels of information.

%% file: sec/method.tex
\section{Janus: A Simple, Unified and Flexible Multimodal Framework}

\begin{figure}[t]
    \centering
    \includegraphics[width=\textwidth]{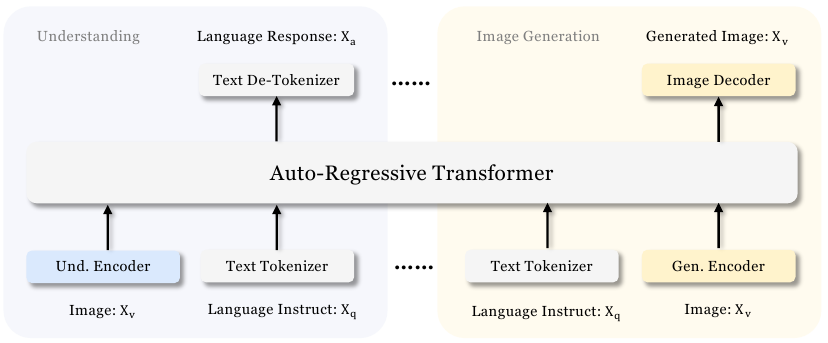}
    \caption{\textbf{Architecture of our Janus.} Different from previous approaches~\cite{team2024chameleon,wu2024vila} that typically assume visual understanding and generation require the same visual encoder, our Janus decouples visual encoding for visual understanding and visual generation. ``Und. Encoder'' and ``Gen. Encoder'' are abbreviations for ``Understanding Encoder'' and ``Generation Encoder'', respectively. Best viewed in color.    } 
    \label{fig:architecture}
\end{figure}

\subsection{Architecture}

The architecture of Janus is shown in Figure~\ref{fig:architecture}. For pure text understanding, multimodal understanding, and visual generation, we apply independent encoding methods to convert the raw inputs into features, which are then processed by an unified autoregressive transformer. Specifically, for text understanding, we use the built-in tokenizer of the LLM to convert the text into discrete IDs and obtain the feature representations corresponding to each ID. For multimodal understanding, we use the SigLIP~\cite{zhai2023sigmoid} encoder to extract high-dimensional semantic features from images. These features are flattened from a $2$-D grid into a $1$-D sequence, and an understanding adaptor is used to map these image features into the input space of the LLM.
For visual generation tasks, we use the VQ tokenizer from~\cite{sun2024autoregressive} to convert images into discrete IDs. After the ID sequence is flattened into $1$-D, we use a generation adaptor to map the codebook embeddings corresponding to each ID into the input space of the LLM. We then concatenate these feature sequences to form a multimodal feature sequence, which is subsequently fed into the LLM for processing. The built-in prediction head of the LLM is utilized for text predictions in both the pure text understanding and multimodal understanding tasks, while a randomly initialized prediction head is used for image predictions in the visual generation task. The entire model adheres to an autoregressive framework without the need for specially designed attention masks.

\subsection{Training Procedure}

\begin{figure}[t]
    \centering
    \includegraphics[width=\textwidth]{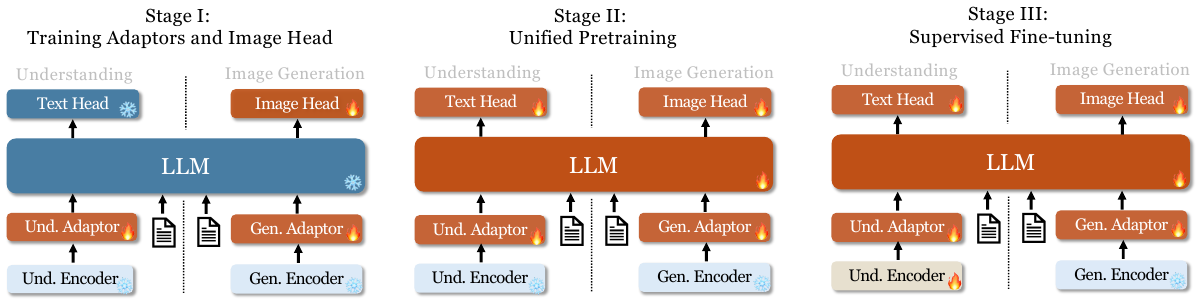}
    \caption{\textbf{Our Janus adopts a three-stage training procedure.} We use flame symbols/snowflake symbols in the diagram to indicate the module updates/does not update its parameters.
    } 
    \label{fig:traing_stages}
\end{figure}

The training of Janus is divided into three stages, as illustrated in Figure~\ref{fig:traing_stages}. Details are provided in the below.

\noindent \textbf{Stage I: Training Adaptors and Image Head.}
The main goal of this stage is to create a conceptual connection between visual and linguistic elements within the embedding space, enabling the LLM to understand the entities shown in images and have preliminary visual generation ability. We keep the visual encoders and the LLM frozen during this stage, allowing only the trainable parameters within the understanding adaptor, generation adaptor and image head to be updated. 

\noindent \textbf{Stage II: Unified Pretraining.}
In this stage, we perform unified pretraining with multimodal corpus to enable Janus to learn both multimodal understanding and generation. 
We unfreeze the LLM and utilize all types of training data: pure text data, multimodal understanding data, and visual generation data. Inspired by Pixart \cite{chen2023pixart}, we begin by conducting simple visual generation training using ImageNet-$1$k to help the model grasp basic pixel dependencies. Subsequently, we enhance the model's open-domain visual generation capability with general text-to-image data.

\noindent \textbf{Stage III: Supervised Fine-tuning.} During this stage, we fine-tune the pretrained model with instruction tuning data to enhance its instruction-following and dialogue capabilities. We fine-tune all parameters except the generation encoder. 
We focus on supervising the answers while masking system and user prompts. To ensure Janus's proficiency in both multimodal understanding and generation, we don't fine-tune separate models for a certain task. Instead, we use a blend of pure text dialogue data, multimodal understanding data and visual generation data, ensuring versatility across various scenarios.

\subsection{Training Objective}
Janus is an autoregressive model, and we simply adopt the cross-entropy loss during training:

\begin{align}
    \mathcal{L} = -\sum_{i=1}\log P_{\theta}(x_i | x_{<i})
\end{align}
Here, $P(\cdot \mid \cdot)$ indicates the conditional probability modeled by the weights $\theta$ of Janus. For pure text understanding and multimodal understanding tasks, we compute the loss on the text sequence. For visual generation tasks, we compute the loss only on the image sequence. To keep the design simple, we have not assigned different loss weights to different tasks.

\subsection{Inference}
During inference, our model adopts a next-token prediction approach. For pure text understanding and multimodal understanding, we follow the standard practice of sampling tokens sequentially from the predicted distribution. For image generation, we utilize classifier-free guidance (CFG)~\footnote{During training, we replace the text condition in the text-to-image data with a pad token at a probability of $10$\%, enabling the model to have unconditional visual generation capability.}, similar to prior works~\cite{gafni2022make,sun2024autoregressive,chang2023muse}. Specifically, for each token, the logit  \(l_g\) is calculated as: \(l_g = l_u + s(l_c - l_u)\), where \(l_c\) is the conditional logit, \(l_u\) is the unconditional logit, and \(s\) is the scale for the classifier-free guidance. The default number of $s$ is $5$ for the following evaluation. 

\subsection{Possible Extensions}
It is important to note that our design, which features separate encoders for understanding and generation, is straightforward and easy to extend.

\noindent \textbf{Multimodal Understanding.} (1) For the multimodal understanding component, a stronger vision encoder can be chosen without worrying about whether the encoder is capable of handling vision generation tasks, such as EVA-CLIP~\cite{sun2023eva}, InternViT~\cite{chen2024internvl}, etc. (2) To handle high-resolution images, dynamic high-resolution techniques~\cite{liu2024improved} can be used. This
allows the model to scale to any resolution, without performing positional embedding interpolation for ViTs. Tokens can be further compressed to save computational cost, for instance, using pixel shuffle operation~\cite{chen2024far}.

\noindent \textbf{Visual Generation.} (1) 
For visual generation, finer-grained encoders can be chosen in order to preserve more image details after encoding, such as MoVQGan~\cite{zheng2022movq}.
(2) Loss functions specifically designed for visual generation can be employed, such as diffusion loss~\cite{li2024autoregressive}.
(3) A combination of AR (causal attention) and parallel (bidirectional attention) methods can be used in the visual generation process to reduce accumulated errors during visual generation~\cite{tian2024visual}.

\noindent \textbf{Support for Additional Modalities.} The straightforward architecture of Janus allows for easy integration with additional encoders, accommodating various modalities such as 3D point cloud \cite{liu2024openshape}, tactile \cite{yang2022touch}, and EEG \cite{bai2023dreamdiffusion}. This gives Janus the potential to become a more powerful multimodal generalist model.

%% file: sec/experiments.tex
\section{Experiments}

In this section, we present a series of comprehensive experiments designed to assess the performance of our method across a range of visual understanding and generation tasks. We begin by detailing our experimental setup, which includes the model architecture, training datasets, and evaluation benchmarks. Next, we report the performance of Janus, followed by a comparison with other state-of-the-art models on various benchmarks for multimodal understanding and generation. We also conduct extensive ablation studies to verify the effectiveness of the proposed method. Lastly, we provide some qualitative results.

\subsection{Implementation Details}

In our experiments, we utilize DeepSeek-LLM ($1.3$B) \cite{bi2024deepseek} with a maximum supported sequence length of $4096$ as the base language model. For the vision encoder used in understanding tasks, we select SigLIP-Large-Patch$16$-$384$ \cite{zhai2023sigmoid}. The generation encoder has a codebook of size $16,384$ and downsamples images by a factor of $16$. Both the understanding adaptor and the generation adaptor are two-layer MLPs. The detailed hyperparameters for each stage are provided in Table~\ref{tab:hyper}. All images are resized to $384 \times 384$ pixels. 
For multimodal understanding data, we resize the long side of the image and pad the short side with the background color (RGB: $127$, $127$, $127$) to reach $384$. 
For visual generation data, the short side is resized to $384$, and the long side is cropped to $384$. We use sequence packing during training to improve training efficiency. We mix all data types according to the specified ratios in a single training step. Our Janus is trained and evaluated using HAI-LLM~\cite{haillm}, which is a lightweight and efficient distributed training framework built on top of PyTorch. The whole training process took $7$ days on a cluster of $16$ nodes, each equipped with $8$ Nvidia A$100$ ($40$GB) GPUs.

\begin{table}[t!]
\centering
\small
\caption{\textbf{Detailed hyperparameters of our Janus}. Data ratio refers to the ratio of multimodal understanding data, pure text data, and visual generation data.}
\label{tab:hyper}
\begin{tabular}{l|ccc}
\toprule
\textbf{Hyperparameters} & \textbf{Stage 1} & \textbf{Stage 2}         & \textbf{Stage 3}  \\
\midrule
Learning rate & $1.0\times10^{-3}$   & $1\times10^{-4}$ & $2.0\times10^{-5}$  \\
LR scheduler  & Cosine & Constant & Constant \\
Weight decay  & $0.0$ & $0.0$ & $0.1$  \\
Gradient clip & $1.0$ & $1.0$ & $1.0$  \\
Optimizer     & \multicolumn{3}{c}{AdamW ($\beta_1=0.9, \beta_2=0.95$)} \\
Warm-up steps    & $300$      & $5,000$  & $0$ \\
Training steps   & $10,000$    & $180,000$ & $24,000$ \\
Batch size       & $256$      & $512$ & $256$   \\
Data Ratio       & $1:0:1$      & $2:3:5$ & $7:3:10$   \\
\bottomrule
\end{tabular}
\end{table}

\begin{table}[ht]
    \centering
    \setlength{\tabcolsep}{1.05pt}
    \renewcommand{\arraystretch}{1.2}
    \scriptsize
    \caption{\textbf{Comparison with state-of-the-arts on multimodal understanding benchmarks}. ``Und.'' and ``Gen.'' denote ``understanding'' and ``generation'', respectively. Models using external pretrained diffusion model are marked with $^\dagger$.}
    \label{sota_result_understanding}
    \begin{tabular}{llccccccccc}
        \toprule
        \textbf{Type} & \textbf{Model} & \textbf{\# LLM Params} & \textbf{POPE$ \uparrow$} & \textbf{MME-P$ \uparrow$} & \textbf{MMB$ \uparrow$} & \textbf{SEED$ \uparrow$} & \textbf{VQAv2$_{(test)}$$\uparrow$} & \textbf{GQA$ \uparrow$} & \textbf{MMMU$ \uparrow$} & \textbf{MM-Vet$ \uparrow$} \\
        \midrule
        \textit{Und. Only} & 
        \cellcolor{blue!3}LLaVA-v$1.5$-Phi-$1.5$~\cite{xie2024show} & \cellcolor{blue!3}$1.3$B & \cellcolor{blue!3}$84.1$ & \cellcolor{blue!3}$1128.0$ & \cellcolor{blue!3}- & \cellcolor{blue!3}- & \cellcolor{blue!3}$75.3$ & \cellcolor{blue!3}$56.5$ & \cellcolor{blue!3}$30.7$ & \cellcolor{blue!3}- \\
        & \cellcolor{blue!3}MobileVLM~\cite{chu2023mobilevlm} & \cellcolor{blue!3}$1.4$B & \cellcolor{blue!3}$84.5$ & \cellcolor{blue!3}$1196.2$ & \cellcolor{blue!3}$53.2$ & \cellcolor{blue!3}- & \cellcolor{blue!3}- & \cellcolor{blue!3}$56.1$ & \cellcolor{blue!3}- & \cellcolor{blue!3}-\\
        & \cellcolor{blue!3}MobileVLM-V2~\cite{chu2024mobilevlm2} & \cellcolor{blue!3}$1.4$B & \cellcolor{blue!3}$84.3$ & \cellcolor{blue!3}$1302.8$ & \cellcolor{blue!3}$57.7$ & \cellcolor{blue!3}- & \cellcolor{blue!3}- & \cellcolor{blue!3}$59.3$ & \cellcolor{blue!3}- & \cellcolor{blue!3}-\\
        
        & MobileVLM~\cite{chu2023mobilevlm} & $2.7$B & $84.9$ & $1288.9$ & $59.6$ & - & - & $59.0$ & - & -\\
        & MobileVLM-V2~\cite{chu2024mobilevlm2} & $2.7$B & $84.7$ & $1440.5$ & $63.2$ & - & - & $61.1$ & - & -\\
        & LLaVA-Phi~\cite{zhu2024llava} & $2.7$B & $85.0$ & $1335.1$ & $59.8$ & - & $71.4$ & - & - & $28.9$\\
        
        & LLaVA~\cite{liu2024visual} & $7$B & $76.3$ & $809.6$ & $38.7$ & $33.5$ & - & - & - & $25.5$ \\
        & LLaVA-v$1.5$~\cite{liu2024improved}& $7$B & $85.9$ & $1510.7$ & $64.3$ & $58.6$ & $78.5$ & $62.0$ & $35.4$ & $31.1$ \\
        & InstructBLIP~\cite{instructblip} & $7$B & - & - & $36.0$ & $53.4$ & - & $49.2$ & - & $26.2$ \\
        & Qwen-VL-Chat~\cite{bai2023qwen} & $7$B & - & $1487.5$ & $60.6$ & $58.2$ & $78.2$ & $57.5$ & - & - \\
        & IDEFICS-$9$B~\cite{laurencon2023introducing} & $8$B & - & - & $48.2$ & - & $50.9$ & $38.4$ & - & - \\
        & Emu$3$-Chat~\cite{wang2024emu3} & $8$B & $85.2$ & - & $58.5$ & $68.2$ & $75.1$ & $60.3$ & $31.6$ & - \\
        & InstructBLIP~\cite{instructblip} & $13$B & $78.9$ & $1212.8$ & - & - & - & $49.5$ & - & $25.6$ \\
        
        \midrule
        \textit{Und. and Gen.} 
        & DreamLLM$^\dagger$~\cite{dong2023dreamllm} & $7$B & - & - & - & - & $72.9$ & - & - & $36.6$ \\
        & LaVIT$^\dagger$~\cite{jin2023unified} & $7$B & - & - & - & - & $66.0$ & $46.8$ & - & - \\
        & Emu$^\dagger$~\cite{sun2023generative} & $13$B & - & - & - & - & $52.0$ & - & - & - \\
        & NExT-GPT$^\dagger$~\cite{wu2023next} & $13$B & - & - & - & - & $66.7$ & - & - & - \\
        \cdashline{2-11}
        \\[-2ex]
        & \cellcolor{blue!3}Show-o~\cite{xie2024show} & \cellcolor{blue!3}$1.3$B & \cellcolor{blue!3}$73.8$ & \cellcolor{blue!3}$948.4$ & \cellcolor{blue!3}- & \cellcolor{blue!3}- & \cellcolor{blue!3}$59.3$ & \cellcolor{blue!3}$48.7$ & \cellcolor{blue!3}$25.1$ & \cellcolor{blue!3}- \\
        & \cellcolor{blue!3}Gemini-Nano-1~\cite{team2023gemini} & \cellcolor{blue!3}$1.8$B & \cellcolor{blue!3}- & \cellcolor{blue!3}- & \cellcolor{blue!3}- & \cellcolor{blue!3}- & \cellcolor{blue!3}$62.7$ & \cellcolor{blue!3}- & \cellcolor{blue!3}$26.3$ & \cellcolor{blue!3}- \\
        & LWM~\cite{liu2024world} & $7$B & $75.2$ & - & - & - & $55.8$ & $44.8$ & - & $9.6$ \\
        & VILA-U~\cite{wu2024vila} & $7$B & $85.8$ & $1401.8$ & - & $59.0$ & $79.4$ & $60.8$ & - & $33.5$ \\
        & Chameleon~\cite{team2024chameleon} & $7$B & - & - & - & - & - & - & $22.4$ & $8.3$ \\
        & \cellcolor{blue!3}\textbf{Janus (Ours)} & \cellcolor{blue!3}$1.3$B & \cellcolor{blue!3}$87.0$ & \cellcolor{blue!3}$1338.0$ & \cellcolor{blue!3}$69.4$ & \cellcolor{blue!3}$63.7$ & \cellcolor{blue!3}$77.3$ & \cellcolor{blue!3}$59.1$ & \cellcolor{blue!3}$30.5$ & \cellcolor{blue!3}$34.3$ \\
        
        \bottomrule
    \end{tabular}
\end{table}

\subsection{Data Setup}
In this section, we provide details of the pretraining and supervised finetuning datasets.

\noindent \textbf{Stage I.} We use a dataset that includes $1.25$ million image-text paired captions from ShareGPT$4$V~\cite{chen2023sharegpt4v} for multimodal understanding and approximately $1.2$ million samples from ImageNet-$1$k~\cite{deng2009imagenet} for visual generation. The ShareGPT$4$V data is formatted as ``$\texttt{<image>}\texttt{<text>}$''. The ImageNet data is organized into a text-to-image data format using the category names: ``$\texttt{<category\_name>}\texttt{<image>}$''. Here, the ``<>'' symbols represent placeholders.

\noindent \textbf{Stage II.} We organize the data into the following categories. (1) Text-only data. We use pretraining text copus from DeepSeek-LLM~\cite{bi2024deepseek}. (2) Interleaved image-text data. We use WikiHow~\cite{koupaee2018wikihow} and WIT~\cite{srinivasan2021wit} dataset. (3) Image caption data. We use images from \cite{deng2009imagenet,kirillov2023segment,dclure_laion_aesthetics_12m_umap,OpenImages,echo840_Detailed_Caption,li2024densefusion,li2024mmsci,li2024multimodal,singla2024pixels}.
Among them, we employ open-source multimodal model to re-caption images in \cite{dclure_laion_aesthetics_12m_umap,OpenImages}. The image caption data is formatted into question-answer pairs, for example, ``$\texttt{<image>} \texttt{Describe the image in detail.} \texttt{<caption>}$''.
 (4) Table and chart data. We use corresponding table and chart data from DeepSeek-VL~\cite{lu2024deepseek}. The data is formatted as ``$\texttt{<question>} \texttt{<answer>}$''. (5) Visual generation data. We utilize image-caption pairs from various datasets including \cite{kirillov2023segment,dclure_laion_aesthetics_12m_umap,OpenImages,singla2024pixels,madebyollin_megalith_10m,mehdidc_yfcc_15m,ProGamerGov_dalle3_1m,pan2023journeydb}, along with $2$M in-house data. For images from \cite{kirillov2023segment,singla2024pixels}, we filter based on aesthetic scores and image sizes, resulting in $20\%$ remaining. During training, we randomly use only the first sentence of a caption with a $25\%$ probability to encourage the model to develop strong generation capabilities for short descriptions. ImageNet samples~\cite{deng2009imagenet} are presented only during the first $120$K training steps, while images from other datasets appear in the later $60$K steps. This approach helps the model first learn basic pixel dependencies before progressing to more complex scene understanding, as suggested by~\cite{chen2023pixart}. The visual generation data is provided in the format: ``$\texttt{<caption>}\texttt{<image>}$''.

\noindent \textbf{Stage III.} For text understanding, we use data from ~\cite{li2024llava}. For multimodal understanding, we use instruct tuning data from~\cite{li2024llava,goyal2017vqav2,hudson2019gqa,shah2019kvqa,hsiao2022screenqa,lu2021iconqa}. For visual generation, we use image-text pairs from \cite{dclure_laion_aesthetics_12m_umap,singla2024pixels,pan2023journeydb} (a subset of that in stage II) and $4$M in-house data. We utilize the following format for instruction tuning:``\texttt{User:}\texttt{<Input Message>} \texttt{\textbackslash n Assistant:} \texttt{<Response>}''. For multi-turn dialogues, we repeat this format to structure the data.
\label{sec:data_setup}

\subsection{Evaluation Setup}

\noindent \textbf{Multimodal Understanding.}
To assess multimodal understanding capabilities, we evaluate our model on widely recognized image-based vision-language benchmarks, which include VQAv2 \cite{goyal2017vqav2}, GQA \cite{hudson2019gqa}, POPE \cite{li2023evaluating}, MME \cite{fu2023mme}, SEED \cite{li2023seed}, MMB \cite{liu2023mmbench}, MM-Vet \cite{yu2023mm}, and MMMU \cite{yue2024mmmu}. 

\noindent \textbf{Visual Generation.}
For evaluating visual generation capabilities, we use the MSCOCO-$30$K \cite{chen2015microsoft}, MJHQ-$30$K \cite{li2024playground}, and GenEval \cite{ghosh2024geneval} benchmarks. MSCOCO-$30$K and MJHQ-$30$K employ the Fréchet Inception Distance (FID) metric on generated images compared to $30$K high-quality images, which indicates the overall efficacy of image generation. GenEval is a challenging benchmark for image-to-text generation, designed to reflect the comprehensive generative abilities of visual generation models by offering a detailed instance-level analysis of their compositional capabilities.
\label{sec:evluation_setup}

\begin{table}[t]
    \centering
    \setlength{\tabcolsep}{4pt}
    \renewcommand{\arraystretch}{1.2}
    \scriptsize
    \caption{\textbf{Evaluation of text-to-image generation ability on GenEval benchmark}. ``Und.'' and ``Gen.'' denote ``understanding'' and ``generation'', respectively. Models using external pretrained diffusion model are marked with $^\dagger$. 
    }
    \begin{tabular}{llcccccccc}
        \toprule
        \textbf{Type} & \textbf{Method} & \textbf{\# Params} & \textbf{Single Obj.} & \textbf{Two Obj.} & \textbf{Counting} & \textbf{Colors} & \textbf{Position} & \textbf{Color Attri.} & \textbf{Overall$\uparrow$} \\
        \midrule
        \multirow{8}{*}{\textit{Gen. Only}} 
        & LlamaGen~\cite{sun2024autoregressive} & $0.8$B & $0.71$ & $0.34$ & $0.21$ & $0.58$ & $0.07$ & $0.04$ & $0.32$ \\
        & LDM~\cite{rombach2022high} & $1.4$B & $0.92$ & $0.29$ & $0.23$ & $0.70$ & $0.02$ & $0.05$ & $0.37$ \\
        & SDv$1.5$~\cite{rombach2022high} & $0.9$B & $0.97$ & $0.38$ & $0.35$ & $0.76$ & $0.04$ & $0.06$ & $0.43$ \\
        & PixArt-$\alpha$~\cite{chen2023pixart} & $0.6$B & $0.98$ & $0.50$ & $0.44$ & $0.80$ & $0.08$ & $0.07$ & $0.48$ \\
        & SDv$2.1$~\cite{rombach2022high} & $0.9$B & $0.98$ & $0.51$ & $0.44$ & $0.85$ & $0.07$ & $0.17$ & $0.50$ \\
        & DALL-E $2$~\cite{ramesh2022hierarchical} & $6.5$B & $0.94$ & $0.66$ & $0.49$ & $0.77$ & $0.10$ & $0.19$ & $0.52$ \\
        & Emu$3$-Gen ~\cite{wang2024emu3} & $8$B & $0.98$ & $0.71$ & $0.34$ & $0.81$ & $0.17$ & $0.21$ & $0.54$ \\
        & SDXL~\cite{podell2023sdxl} & $2.6$B & $0.98$ & $0.74$ & $0.39$ & $0.85$ & $0.15$ & $0.23$ & $0.55$ \\
        \midrule
        \multirow{5}{*}{\textit{Und. and Gen.}}
        & SEED-X$^\dagger$~\cite{ge2024seed} & $17$B & $0.97$ & $0.58$ & $0.26$ & $0.80$ & $0.19$ & $0.14$ & $0.49$ \\
        \cdashline{2-10}
        & Show-o~\cite{xie2024show} & $1.3$B & $0.95$ & $0.52$ & $0.49$ & $0.82$ & $0.11$ & $0.28$ & $0.53$ \\
        & LWM~\cite{liu2024world} & $7$B & $0.93$ & $0.41$ & $0.46$ & $0.79$ & $0.09$ & $0.15$ & $0.47$ \\
        & Chameleon~\cite{team2024chameleon} & $34$B & - & - & - & - & - & - & $0.39$ \\
        & \textbf{Janus (Ours)} & $1.3$B & $0.97$ & $0.68$ & $0.30$ & $0.84$ & $0.46$ & $0.42$ & $0.61$ \\
        
        \bottomrule
    \end{tabular}
    \label{tab:geneval}
\end{table}

\subsection{Comparison with State-of-the-arts}
\noindent \textbf{Multimodal Understanding Performance.}
We compare the proposed method with state-of-the-art unified models and understanding-only models in Table~\ref{sota_result_understanding}.
Janus achieves the overall best results among models of similar scale. Specifically, compared to the previous best unified model, Show-o~\cite{xie2024show}, we achieve performance improvements of $41$\% ($949 \rightarrow 1338$) and $30$\% ($48.7 \rightarrow 59.1$) on the MME and GQA datasets, respectively. This can be attributed to Janus decoupling the visual encoding for multimodal understanding and generation, mitigating the conflict between these two tasks.
When compared to models with significantly larger sizes,  Janus remains highly competitive. For instance, Janus outperforms LLaVA-v$1.5$ ($7$B) on several datasets, including POPE, MMbench, SEED Bench, and MM-Vet.

\noindent \textbf{Visual Generation Performance.}
We report visual generation performance on GenEval, COCO-$30$K and MJHQ-$30$K benchmarks. As shown in Table~\ref{tab:geneval}, our Janus obtains $61$\% overall accuracy on GenEval, which outperforms the previous best unified model Show-o ($53$\%) and some popular generation-only methods, e.g., SDXL ($55$\%) and DALL-E $2$ ($52$\%). This demonstrates that our approach has better instruction-following capabilities. As shown in Table~\ref{tab:fid}, Janus achieves FIDs of $8.53$ and $10.10$ on the COCO-$30$K and MJHQ-$30$K benchmarks, respectively, surpassing unified models Show-o and LWM, and demonstrating competitive performance compared to some well-known generation-only methods. This demonstrates that the images generated by Janus have good quality and highlights its potential in visual generation.

\begin{table}[ht]
    \centering
    \scriptsize
    \setlength{\tabcolsep}{3pt}
    \renewcommand{\arraystretch}{1.2}
    \caption{\textbf{Evaluation of text-to-image generation ability on MSCOCO-$30$K and MJHQ-$30$K benchmark}. ``Und.'' and ``Gen.'' denote ``understanding'' and ``generation'', respectively. Models using external pretrained diffusion model are marked with $^\dagger$.  
    }
    \begin{tabular}{lp{3.5cm}ccr}
        \toprule
        \textbf{Type} & \textbf{Model} & \textbf{\# Params} & \textbf{COCO-30K$\downarrow$} & \textbf{MJHQ-30K$\downarrow$} \\
        \midrule
        \multirow{9}{*}{\textit{Gen. Only}}
        & DALL·E~\cite{ramesh2021zero} & $12$B & $27.50$ & - \\
        & GLIDE~\cite{nichol2021glide} & $5$B & $12.24$ & - \\
        & LDM~\cite{rombach2022high} & $1.4$B & $12.64$ & - \\
        & DALL·E 2~\cite{ramesh2022hierarchical} & $6.5$B & $10.39$ & - \\
        & SDv1.5~\cite{rombach2022high} & $0.9$B & $9.62$ & - \\
        & GigaGAN~\cite{kang2023scaling} & $0.9$B & $9.09$ & - \\
        & PixArt-$\alpha$~\cite{chen2023pixart} & $0.6$B & $7.32$ & - \\
        & Imagen~\cite{saharia2022photorealistic} & $34$B & $7.27$ & - \\
        & RAPHAEL~\cite{xue2024raphael} & $3$B & $6.61$ & - \\
        \midrule
        \multirow{4}{*}{\textit{Und. and Gen.}} 
        & Emu$^\dagger$~\cite{sun2023generative} & $13$B & $11.66$ & - \\
        & NExT-GPT$^\dagger$~~\cite{wu2023next} & $13$B & $11.28$ & - \\
        & SEED-X$^\dagger$~\cite{ge2024seed} & $17$B & $14.99$ & - \\
        \cdashline{2-5}
        
        & Show-o~\cite{xie2024show} & $1.3$B & $9.24$ & $15.18$ \\
        & LWM~\cite{liu2024world} & $7$B & $12.68$ & $17.77$ \\
        & VILA-U ($256$)~\cite{wu2024vila} & $7$B & - & $12.81$ \\
        & VILA-U ($384$)~\cite{wu2024vila} & $7$B & - & $7.69$ \\
        & \textbf{Janus (Ours)} & $1.3$B & $8.53$ & $10.10$ \\
        \bottomrule
    \end{tabular}
    \label{tab:fid}
\end{table}

\begin{table}[t]
\centering
\caption{\textbf{Ablation studies}. We verify the effectiveness of decoupling visual encoding and compare unified training with task-specific training. ``Und.'', ``Gen.'' and ``SE. Tokenizer'' denote ``understanding'', ``generation'' and ``semantic tokenizer'', respectively. 
}
\setlength{\tabcolsep}{4pt}
\renewcommand{\arraystretch}{1.2}
\small
\begin{tabular}{cccccccc}
\toprule
\textbf{Exp ID} & \textbf{Visual Encoder} &  \textbf{Training Task} & \textbf{POPE}$\uparrow$ & \textbf{MMB}$\uparrow$ & \textbf{SEED}$\uparrow$ & \textbf{MMMU}$\uparrow$ & \textbf{COCO-FID}$\downarrow$ \\ 
\midrule
A & VQ Tokenizer        & Und. + Gen.      & $60.1$  & $35.0$   & $34.9$   & $24.7$  & $8.72$      \\ 
B & SE. Tokenizer  & Und. + Gen.       & $82.4$  & $52.7$   & $54.9$   & $26.6$  & $7.11$      \\ 
C & SE. Tokenizer           & Und.  & $83.9$  & $62.1$ & $60.8$  & $27.5$  & -        \\ 
D & SigLIP + VQ (Ours)            & Und. + Gen.    & $87.0$ & $69.4$  & $63.7$  & $30.5$  &  $8.53$   \\ 
\midrule
E & SigLIP            & Und.  & $85.9$  & $70.6$ & $64.8$  & $28.8$  & -        \\ 
F & VQ Tokenizer           & Gen.       & -    & -   & -    & -   & $8.92$        \\ 
\bottomrule
\end{tabular}
\label{tab:ablation}
\end{table}

\subsection{Ablation Studies}

We carefully design ablation studies to verify the effectiveness of Janus's design concept. First, we design experiments to validate the importance and benefits of decoupling visual encoding. Second, we investigate the impact of unified training on individual tasks like multimodal understanding or visual generation. Results are listed in Table~\ref{tab:ablation}.

\noindent \textbf{Baseline Construction.} Following previous work~\cite{team2024chameleon}, we select a VQ tokenizer~\cite{sun2024autoregressive} to encode images for both multimodal understanding and generation tasks, serving as the baseline (Exp-A). Considering that the VQ tokenizer in Exp-A might be weak in extract semantic information, making it less effective for multimodal understanding, we also construct a stronger baseline Exp-B. We adopt SigLIP to distill an enhanced semantic tokenizer~\footnote{The semantic tokenizer is only used in the ablation study as a stronger baseline. For simplicity, we use the ordinary VQ tokenizer~\cite{sun2024autoregressive} in the main experiment.} that can extract high-level semantic information from images while also have the ability to convert images into discrete IDs, which is similar to that in~\cite{wu2024vila}. Details of the semantic tokenizer could be found in the Appendix ~\ref{sec:se_dec_appendix}.

\noindent \textbf{Impact of Decoupling Visual Encoding.} 
(1) From the results of Exp-A, we find the model achieves satisfactory performance on visual generation benchmark ($8.72$ FID on COCO). However, there is a significant gap on understanding benchmarks between Exp-A and our model (Exp-D). 
(2) When comparing Exp-B to Exp-A, the results show a clear improvement in multimodal understanding, although there is still a considerable gap compared to our method. In terms of visual generation, Exp-B outperforms Exp-D. We hypothesize two possible reasons for this. First, the semantic tokenizer produces discrete IDs that are more semantically coherent, providing more reasonable prediction targets for the LLM. Second, the visual encoder in Exp-B has significantly more parameters than the Gen. encoder in Exp-D. 
(3) To investigate whether using a single visual encoder leads to a trade-off between multimodal understanding and generation, we further design Exp-C based on Exp-B, which focuses solely on multimodal understanding training. The multimodal understanding ability of Exp-C is significantly better than that of Exp-B. This indicates that the visual encoder in Exp-B made trade-offs between multimodal understanding and generation, ultimately sacrificing its multimodal understanding capability. The above experiments illustrate the importance of decoupling visual encoding.

\noindent \textbf{Unified Model vs. Pure Understanding \& Pure Generation.} We compare the performance of unified training (Exp-D) against pure understanding (Exp-E) and pure generation (Exp-F) training. For pure understanding, we omit visual generation data. For pure generation, we exclude the understanding data. 
Please note that unified training and pure understanding training go through the same steps for the understanding part. Similarly, unified training and pure generation training go through the same steps for the visual generation part.
Experimental results show that the performance of unified training is comparable to that of training solely for understanding or solely for visual generation. This demonstrates that our model, Janus, is capable of incorporating strong generative abilities while minimally affecting multimodal understanding performance.

\begin{figure}[t!]
    \centering
    \includegraphics[width=\textwidth]{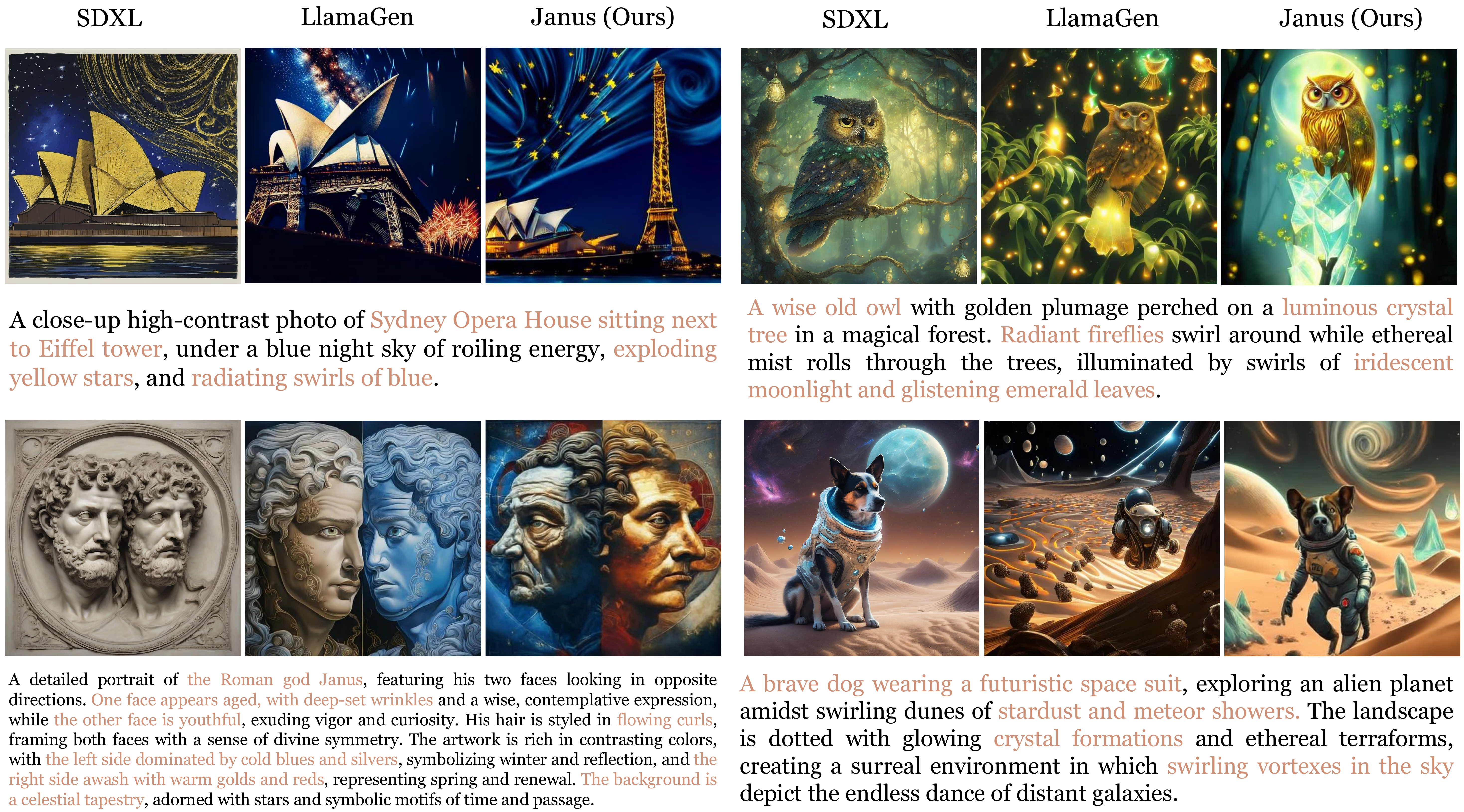}
    \caption{\textbf{Qualitative comparisons of visual generation with LlamaGen and SDXL}. The images generated by Janus show better consistency with the user's prompts. The image resolutions for SDXL, LlamaGen, and ours are $1024 \times 1024$, $512 \times 512$, and $384 \times 384$, respectively. Best viewed on screen.
    } 
    \label{fig:gen_sample}
\end{figure}

\subsection{Qualitative Results}

\begin{figure}[ht!]
    \centering
    \includegraphics[width=\textwidth]{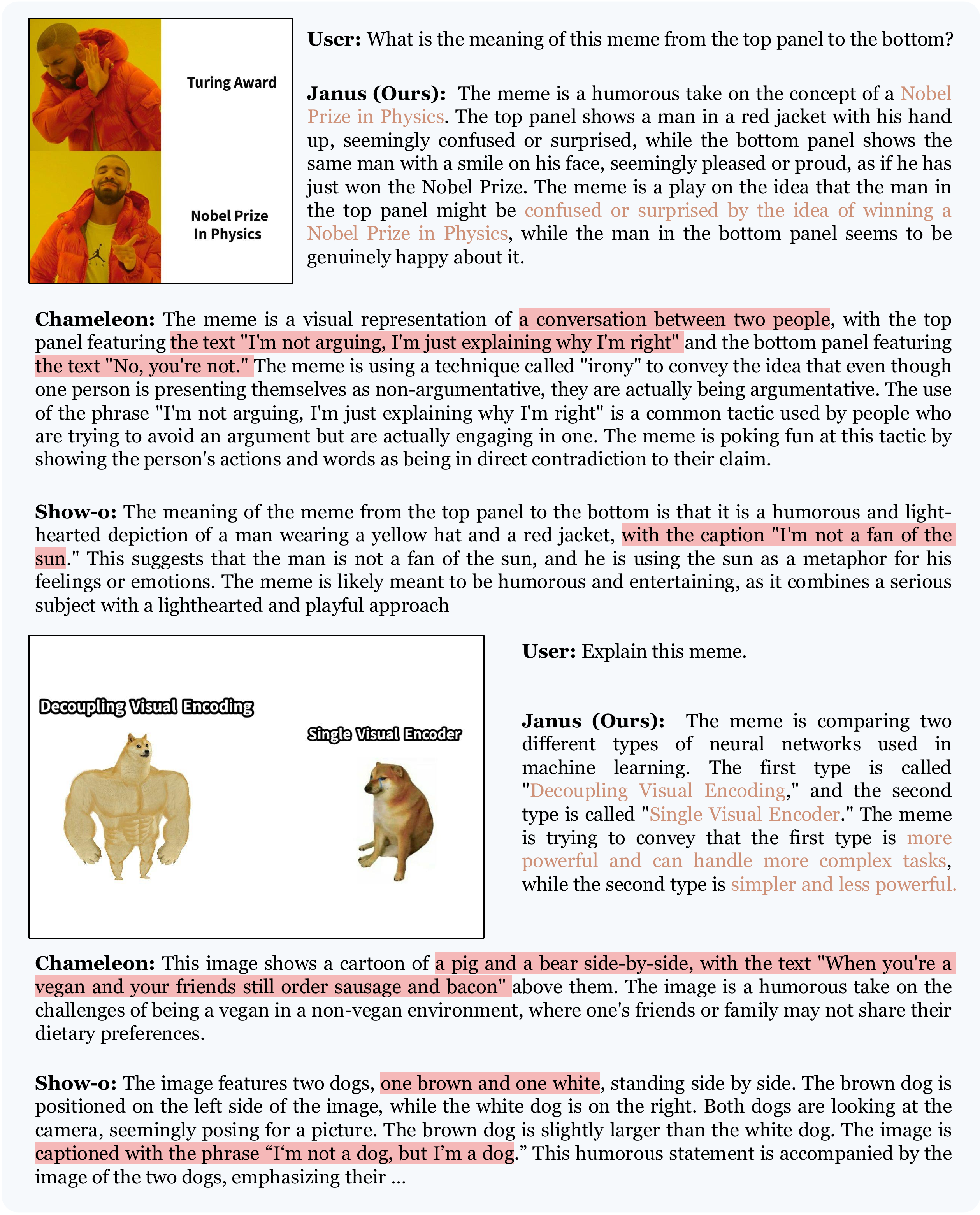}
    \caption{\textbf{Qualitative results of multimodal understanding on humorous memes}. We compare the response with Chameleon-$7$B~\cite{team2024chameleon} and Show-o~\cite{xie2024show}. We emphasize the key-points in the response. Best viewed on screen.
    } 
    \label{fig:mm_und_sample}
\end{figure}

\noindent \textbf{Visualizations of Visual Generation.} Figure \ref{fig:gen_sample} provides qualitative comparisons between our model, diffusion-based models like SDXL~\cite{podell2023sdxl}, and the autoregressive model LlamaGen~\cite{sun2024autoregressive}.  The results show that our model demonstrates superior instruction-following capabilities in visual generation, accurately capturing most of details in the user’s prompt. This indicates the potential of the unified model in the realm of visual generation. More visualizations can be found in the Appendix ~\ref{sec:appendix_qualitative}.

\noindent \textbf{Multimodal Understanding on MEME Images.} Figure~\ref{fig:mm_und_sample} showcases the qualitative results of Janus's multimodal understanding ability, compared with Chameleon~\cite{team2024chameleon} and Show-o~\cite{xie2024show}. Janus accurately interprets the text caption and captures the emotion conveyed in the meme. In contrast, both Chameleon and Show-o struggle with accurately recognizing the text in the image. Additionally, Chameleon fails to identify objects in the meme, while Show-o misinterprets the dog's color. These examples highlight that the decoupled vision encoder significantly enhances Janus's fine-grained multimodal understanding ability compared to the shared encoder used by Chameleon and Show-o. More multimodal understanding exmples can be found in the Appendix ~\ref{sec:appendix_qualitative}.

%% file: sec/conclusion.tex
\section{Conclusion}

In this paper, we introduced Janus, a simple, unified and extensible multimodal understanding and generation model. The core idea of Janus is to decouple visual encoding for multimodal understanding and generation, which could alleviate the conflict arising from the differing demands that understanding and generation place on the visual encoder. Extensive experiments have demonstrated the effectiveness and leading performance of Janus. It is also worth noting that Janus is flexible and easy to extend. In addition to having significant potential for improvement in both multimodal understanding and generation, Janus is also easily extendable to incorporate more input modalities. The above advantages suggest that Janus may serve as an inspiration for the development of the next generation of multimodal general-purpose models.

%% file: sec/appendix.tex
\appendix
\section*{Appendix}

\section{Details of Semantic Tokenizer Mentioned in Ablation Study}
\subsection{Architecture of Semantic Tokenizer}
~\label{sec:se_dec_appendix}

\begin{figure}[ht]
    \centering
    \includegraphics[width=\textwidth]{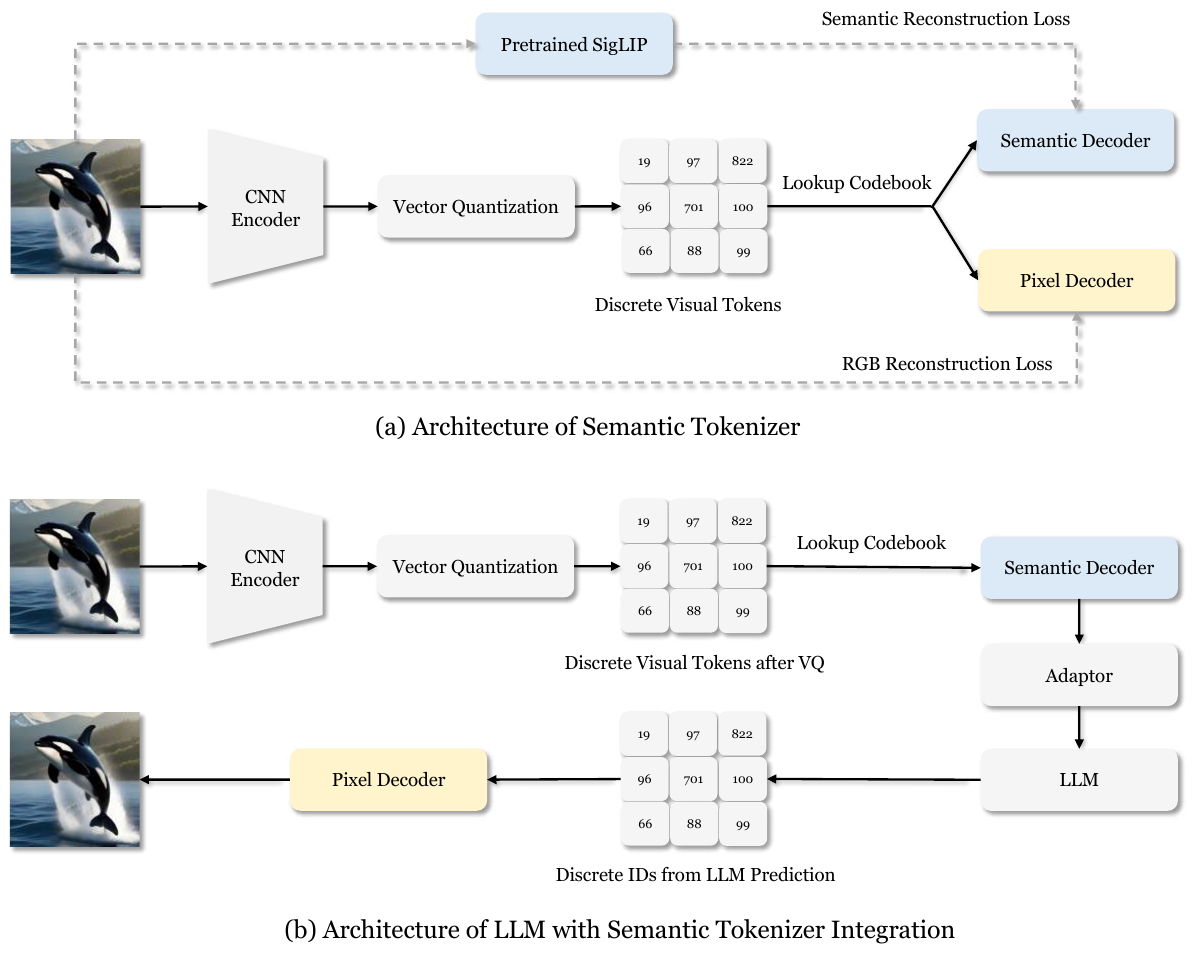}
    \caption{\textbf{Architecture and usage of the semantic tokenizer.} (a) Architecture used during training of the semantic tokenizer. We use pre-trained SigLIP~\cite{zhai2023sigmoid} to supervise the reconstruction of semantic information, while using raw image to supervise the reconstruction of RGB values.  (b) Integrating LLM with the semantic decoder. The semantic decoder outputs continuous features with high-level semantics, which are passed through an adaptor and then used as input for the LLM. Please note that the semantic tokenizer is only used in the ablation study, not in the main experiment.
    } 
    \label{fig:semantic_tokenizer}
\end{figure}

We build the semantic tokenizer based on the tokenizer architecture proposed in~\cite{sun2024autoregressive}, which has a downsample rate of $16$. In addition to the original CNN pixel decoder, we add an additional semantic decoder branch after Vector Quantization, as shown in Figure~\ref{fig:semantic_tokenizer} (a). The semantic decoder is a $12$-layer ViT~\cite{dosovitskiy2020image}, with $12$ attention heads and a hidden dimension of $768$. For the semantic decoder, we use a causal attention mask to facilitate next token prediction when integrating it with an LLM.

\subsection{Training}
\noindent \textbf{Training Procedure.} 
The semantic tokenizer is trained from scratch in a two-stage manner. In the first stage, we train the model on the ImageNet-$1$k~\cite{deng2009imagenet} dataset for $40$ epochs. In the second stage, we fine-tune the model for $1$ epoch on $50$ million images. These images come from the visual generation data used during the Janus pretraining process. We use a constant learning rate of $1e-4$ and a batch size of $128$.

\noindent \textbf{Training Loss.}
The training loss of the semantic tokenizer consists of two parts. On one hand, we use the loss for RGB reconstruction as described in \cite{sun2024autoregressive}. On the other hand, we use SigLIP-Large-Patch$16$-$384$ as the teacher to supervise the semantic feature reconstruction results by the semantic decoder.
We adopt the loss in BEiT-v2~\cite{peng2022beit}. Specifically, we maximize the cosine similarity between the semantic feature predicted by the semantic decoder and the SigLIP output. The weight for the semantic reconstruction loss is set to $0.25$.

\subsection{Integrating with LLM}
We present the integration of the semantic tokenizer and the LLM in Figure~\ref{fig:semantic_tokenizer} (b). The image is first transformed into continuous features through the CNN encoder, vector quantization and the semantic decoder. Then, the LLM processes these features and generates predictions for the image IDs. Finally, the pixel decoder converts these discrete IDs into RGB values.

\section{Additional Qualitative Results}

\begin{figure}[ht!]
    \centering
    \includegraphics[width=0.84\textwidth]{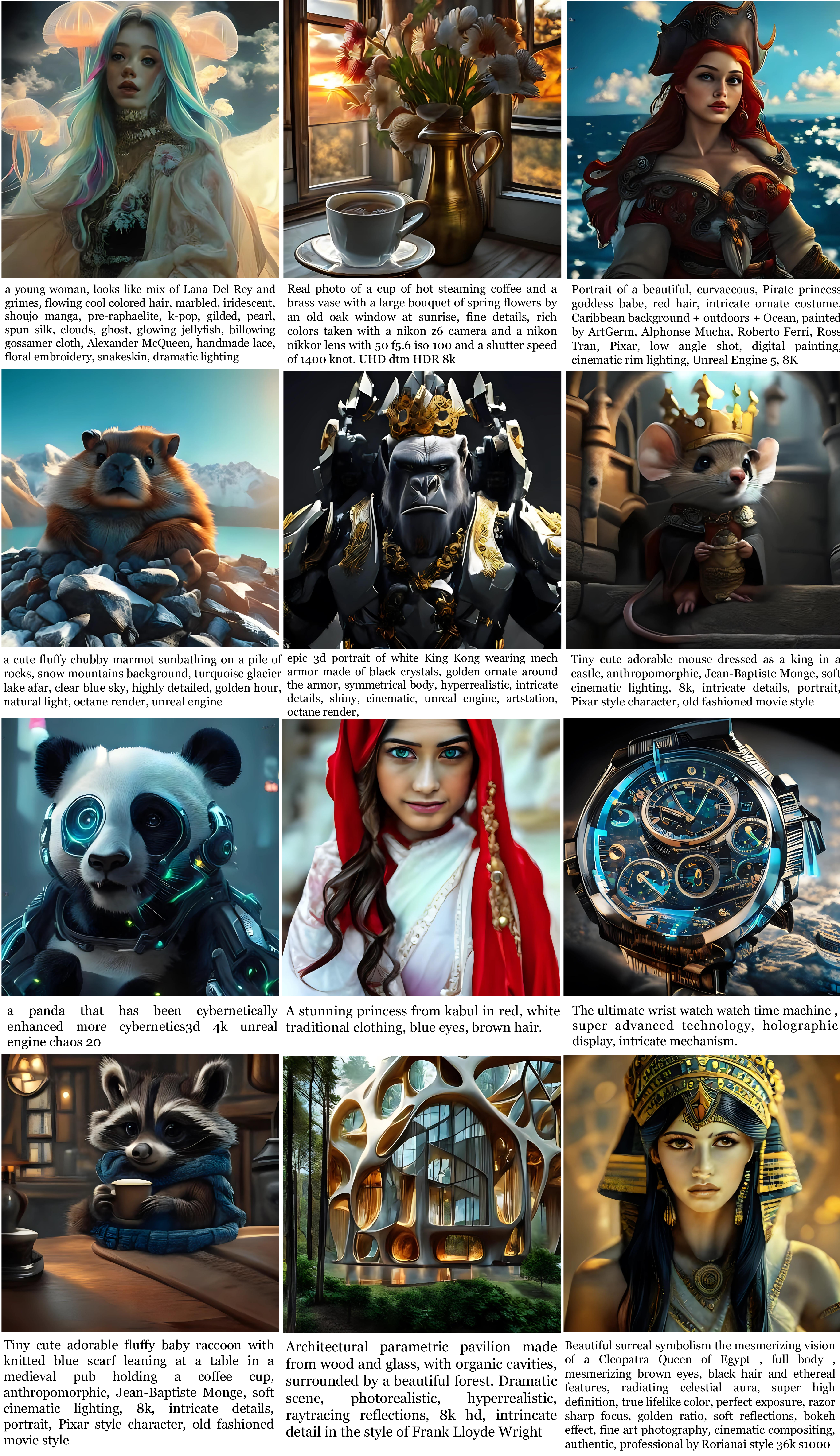}
    \caption{More text-to-image generation results. We upsample the images to $1024 \times 1024$ for better visualization.} 
    \label{fig:t2i_appendix}
\end{figure}

\begin{figure}[h!]
    \centering
    \includegraphics[width=0.85\textwidth]{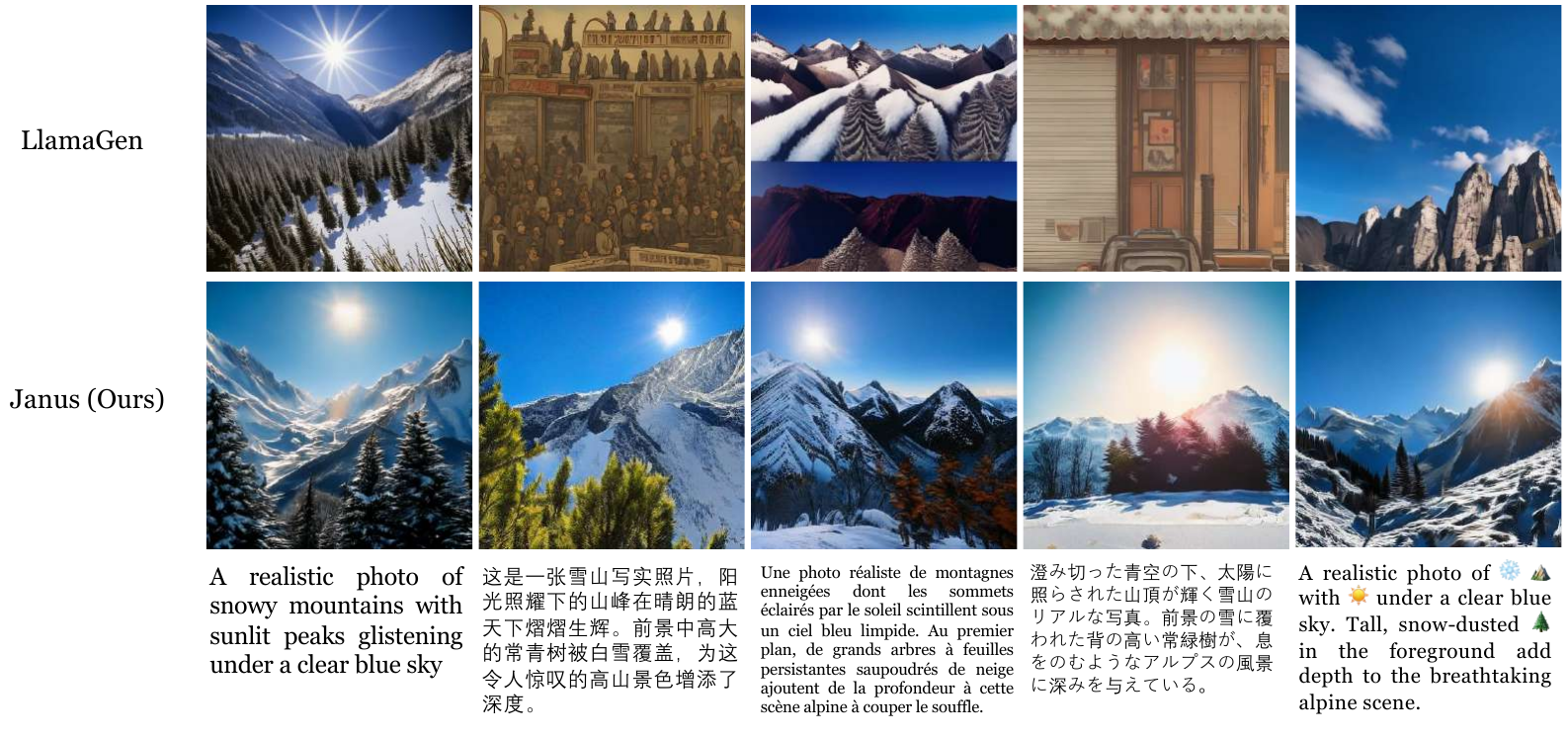}
    \vspace{-3mm}
    \caption{Multilingual text-to-image generation samples compared to LlamaGen~\cite{sun2024autoregressive}. Note that we only use English text-to-image data in training, and this is an emergent capability of our model. The languages used in the prompt, from left to right, are: English, Chinese, French, Japanese, and English with emoji.} 
    \label{fig:multilingual_t2i_appendix}
\end{figure}

\noindent \textbf{More Visualizations of Text-to-Image Generation.} We present more text-to-image generation results in Figure~\ref{fig:t2i_appendix}. It is evident that Janus is capable of producing high-quality images that adhere closely to the given prompts. We further explore the multilingual text-to-image capabilities of our model, as shown in Figure~\ref{fig:multilingual_t2i_appendix}. We are pleasantly surprised to find that, despite our training data consisting solely of English text-to-image data, Janus can still process text-to-image tasks in other languages. We attribute this multilingual ability to the original large language model's inherent traits. The LLM initially translates various languages into a unified semantic space, allowing Janus to perform text-to-image tasks naturally without additional training.

\begin{figure}[h!]
    \centering
    \includegraphics[width=0.7\textwidth]{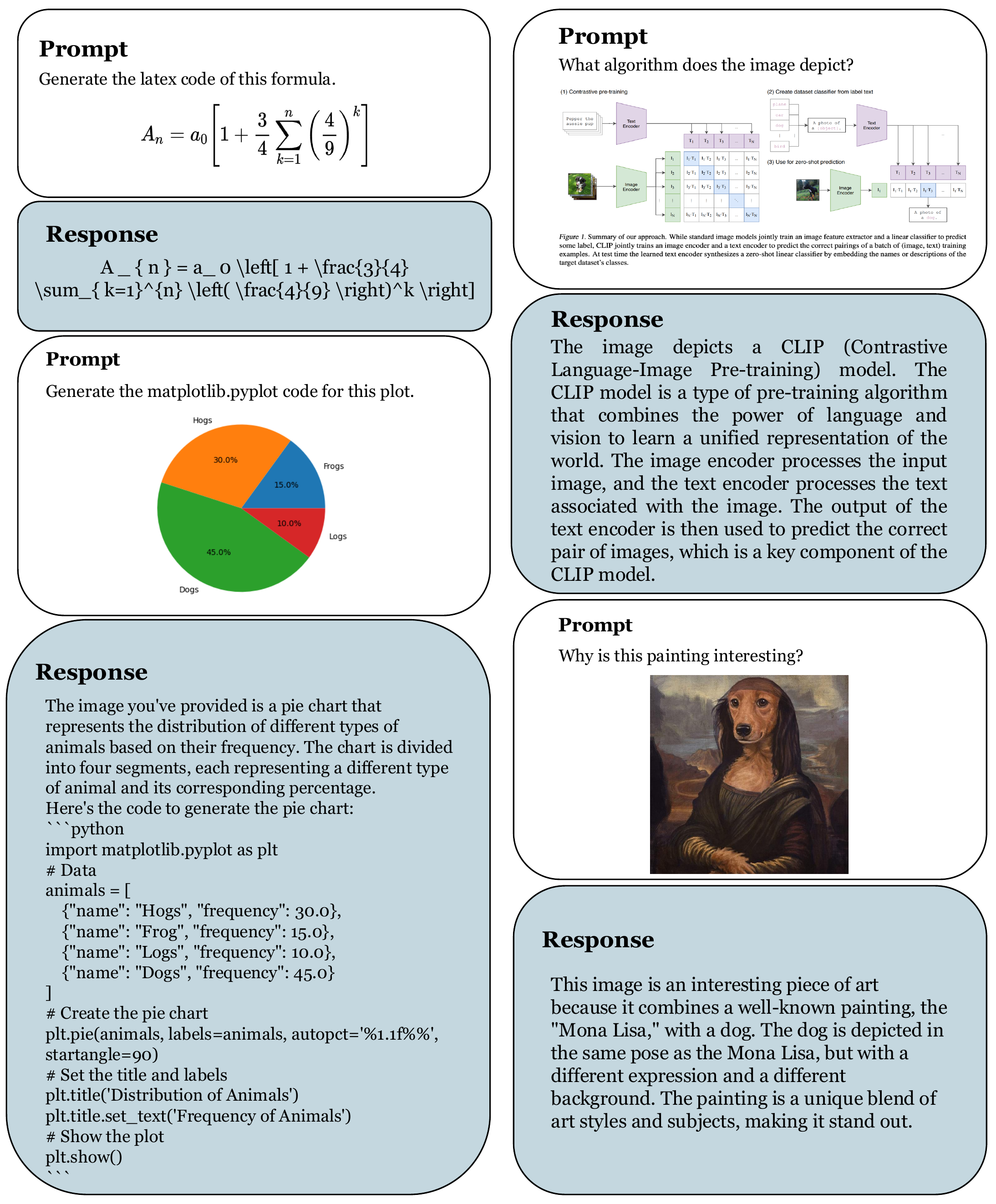}
    \vspace{-3mm}
    \caption{More multimodal understanding results. Janus has a strong multimodal understanding capability and can handle inputs from various contexts, such as scientific charts, artwork images, LaTeX formula images, and more.} 
    \label{fig:mm_und_appendix}
\end{figure}

\noindent \textbf{More Multimodal Understanding Results.} Additional results on multimodal understanding are shown in Figure~\ref{fig:mm_und_appendix}. Janus exhibits impressive comprehension abilities when handling inputs from various contexts, showcasing its powerful capabilities.

\label{sec:appendix_qualitative}

